\documentclass[10pt,twocolumn,letterpaper]{article}

\usepackage{iccv}
\usepackage{times}
\usepackage{epsfig}
\usepackage{graphicx}
\usepackage{amsmath}
\usepackage{amssymb}

\usepackage[breaklinks=true,bookmarks=false]{hyperref}

\iccvfinalcopy 

\def\iccvPaperID{****} 

\ificcvfinal\fi

\begin{document}

\title{WAD: A Deep Reinforcement Learning Agent for Urban Autonomous Driving}

\author{Arjit Sharma\\
Liverpool John Moores University\\
Liverpool, United Kingdom\\
{\tt\small sharma.arjit07@gmail.com}
\and
Sahil Sharma\\
Thapar Institute of Engineering and Technology\\
Patiala, India\\
{\tt\small sahil301290@gmail.com}
}

\maketitle

\ificcvfinal\thispagestyle{empty}\fi


\begin{abstract}

   Urban autonomous driving is an open and challenging problem to solve as the decision-making system has to account for several dynamic factors – like multi-agent interactions, diverse scene perceptions, complex road geometries, and other rarely occurring real-world events. On the other side, with deep reinforcement learning (DRL) techniques, agents have learned many complex policies. They have even achieved super-human-level performances in various Atari Games and Deepmind's AlphaGo. However, current DRL techniques do not generalize well on complex urban driving scenarios. This paper introduces the DRL driven Watch and Drive (WAD) agent for end-to-end urban autonomous driving. Motivated by recent advancements, the study aims to detect important objects/states in high dimensional spaces of CARLA and extract the latent state from them. Further, passing on the latent state information to WAD agents based on TD3 and SAC methods to learn the optimal driving policy. Our novel approach utilizing fewer resources, step-by-step learning of different driving tasks, hard episode termination policy, and reward mechanism has led our agents to achieve a 100\% success rate on all driving tasks in the original CARLA benchmark and set a new record of 82\% on further complex NoCrash benchmark, outperforming the state-of-the-art model by more than +30\% on NoCrash benchmark.
\end{abstract}

\section{Introduction}

An intelligent and context-aware decision-making system is vital for urban autonomous driving. The driving agent should be able to follow lanes, detect vehicles, pedestrians, traffic signals, and should be able to respond to some rare events of road construction, a pedestrian suddenly crossing the road, or even an accident ahead. This decision-making problem becomes even more challenging when all the dynamic factors have to be dealt with at once. Learning to autonomously drive in high dimensional and dynamic environments with classic rules-based approaches~\cite{r1, r29}, - which manually design the driving policy, becomes complicated. Furthermore, these rule-based policies become biased to the training environment and require manual redesigning of policy under different tasks and scenarios.

\begin{figure*}[htbp]
    
       \includegraphics[width=1.0\linewidth]{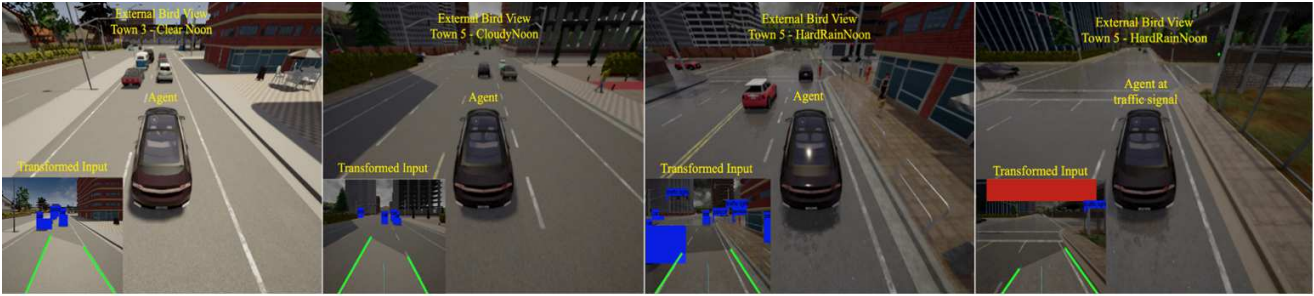}
       \caption{WAD agent driving in the urban environment of CARLA 0.9.6 simulation. Figure shows the external bird view of agent driving under different weather conditions in medium to dense traffic and the transformed input image with object detection, lane, and traffic light detection done in the bottom left corner image.}
    \label{fig:long}
    \end{figure*}

Recent state-of-the-art techniques in autonomous driving~\cite{10.1007/978-3-030-01234-2_36, r3} have focused on learning driving policy directly from data without relying much on handwritten rules. Two techniques majorly studied for learning end-to-end driving policy are - Imitation Learning (IL) and Reinforcement Learning (RL).

IL technique aims to learn the driving policy by imitating controls applied by the human expert while driving the vehicle~\cite{r4,r5}. It requires large amounts of synthesized driving data to learn the policy. With some shortcomings to it~\cite{r6}, policies are learned mostly on perfect driving data, with less data on dangerous accidental situations; these policies almost do not react appropriately under accidental situations as they suffer from distribution mismatch. Moreover, it remains costly and time-consuming to collect and process vast amounts of expert driving data in the real world and in real-time to learn the optimal policy.

RL~\cite{r32}, combined with deep learning techniques, on the other side, lets the agent learn the policy on its own by providing reward signals from the environment for the action it has taken at each timestep. Based on these reward signals, the RL agent determines the appropriateness of the action taken and keeps improving its actions, thereby formulating the optimal driving policy. This learning type does not suffer from a distribution mismatch. It does not require any human guidance, saving both time and human resources, making them a good alternative over IL techniques for urban autonomous driving.

Recent advancements in DRL techniques~\cite{r33, r34} have also shown agents achieving super-human-level performances in Atari games~\cite{r1}, Deepmind's AlphaGo~\cite{r8}, and even in car racing games like TORCS~\cite{r9}. The progress has primarily been because the agent is trained to do a single task or perform multiple tasks at different game stages. These games do not present complexities of urban driving like diverse road geometries, cross intersections, traffic signals, pedestrians, and changing weather conditions.

Further, with high dimensional spaces, DRL techniques face one shortcoming. These techniques rely on replay buffer – memory maintaining past experiences of the agent. It becomes hard to maintain experiences into the agent's memory with high dimensional environments and large image sizes. Even if we allocate a large memory for our agent, the training over these experiences becomes difficult and time-consuming. So, how do we improve upon this shortcoming and use DRL for high-dimensional urban autonomous driving?

One way is instead of directly using images as input and predicting control commands, we can pre-train an additional convolutional neural network (CNN) to predict the low dimensional intermediate representation of the environment as ''Affordances''~\cite{r10}. These affordances, along with other control states, can then be passed to RL agent to perform training and learn the policy~\cite{r17, r18}.

However, this approach's problem is – traffic lights, traffic signs, distant pedestrians, or an approaching child all have tiny pixel region in the overall image. Hence, there remains a high chance of losing out on this essential intermediate information upon encoding and decoding. Secondly, different types of vehicles (bike, car, bus, truck), pedestrians in different clothes, buildings, and various signals all have different shapes, sizes, and even colours. There becomes a high possibility of either misinterpreting this information or missing out on it either due to less training data or improper encoding and decoding. When good quality latent state is not presented to the RL agent, then the training process takes more time, and the driving policy's optimality is impacted.

Another way of dealing with this is, introducing a separate detection module before latent state calculation. This detection module would detect vehicles, pedestrians, traffic light signals, and lanes. Having had these objects/states detected and highlighted would help to extract improved quality latent state from images. Furthermore, the RL agent will learn better policy when better quality latent state is presented to it.

We introduce the DRL driven Watch and Drive (WAD) agent to operate autonomously in Car Learning to Act (CARLA)~\cite{r16} urban driving environment. The trained WAD agent will follow lanes, avoid collisions, and follow traffic light signals. The key contributions can be summarized as: • DRL driven Watch and Drive (WAD) agent based on TD3 and SAC methods will follow the lane, avoid collisions, cross intersections, and obey traffic light signals. • A novel approach for object segmentation by pre-detecting and coloring critical objects/states in the environment • Step-by-step learning of different driving tasks, hard episode termination policy, dense reward mechanism leading to 100\% success rate on CARLA, and 82\% on NoCrash benchmarks, outperforming state-of- the-art models.

\section{Related Work}

The recent progress in DRL techniques combined with the fact that RL techniques rely on trial and error and do not require any supervised data for training has attracted many researchers to try these techniques in the field of autonomous driving.

Karavolos~\cite{Karavolos2013QlearningWH} applied the q-learning technique on TORCS~\cite{r9} simulator and evaluated the heuristic effectiveness during the exploration. Huval et al.~\cite{huval2015empirical} proposed a CNN-based model to detect vehicles and lanes for evaluating their approach on the real-world highway. Bojarski et al.~\cite{r4} proposed an end-to-end model for autonomous driving and tested it on simulators and real- world environments. Wolf et al.~\cite{P.Wolf} used DQN to develop a policy that could provide discrete steer controls to drive a car in simulation. Sallab et al.~\cite{r14} used the DQN technique for discrete actions and deep deterministic actor-critic (DDAC) algorithm for continuous actions to train the agent to assist in lane-keeping using the TORCS simulator. Wang et al.~\cite{r15} used the Q-learning algorithm to train the vehicle for automated lane-change manoeuvres.

Recent studies utilized CARLA~\cite{r16} simulator for urban autonomous driving. The research group of CARLA used IL and RL techniques while releasing a driving benchmark. The RL technique used asynchronous advantage actor-critic algorithm with discrete actions for learning the driving policy. The results of the RL technique were inferior compared to the IL technique. Later, Liang et al.~\cite{10.1007/978-3-030-01234-2_36} built upon supervised learning in a controlled imitation stage, followed by sharing learned weights into the RL stage utilizing deep deterministic policy gradients (DDPG) methods. Following this mixed IL and RL approach, it led them to surpass baseline methods with good margins.

On the other side, Chen et al.~\cite{r10} first proposed to use the CNN model to predict crucial perception indicators that directly relate to the affordance of the road/traffic state for driving and coined the direct perception approach for autonomous driving. They trained the CNN to predict compact yet complete high-level information like – distance to other vehicles, distance to the lane marking, heading angle of the agent within the TORCS simulation environment. These computed affordances were then passed to the rule-based controller for driving. However, their approach was restricted to simple highway driving, and they could not navigate through intersections and follow traffic signals. Sauer et al.~\cite{r23} used the same approach in the complex urban environment of CARLA. They demonstrated to have handled smooth car following, traffic lights, and speed signs - achieving a 68\% success rate on the CARLA benchmark. Both studies used rule-based controllers to make driving decisions.

Chen et al.~\cite{r17} used the variational autoencoder (VAE) technique~\cite{r35, r36} to extract low-level information from the bird-eye-view image of CARLA to be passed through various RL algorithms for learning the driving policy. However, their study was limited to a dense traffic roundabout scenario. Recently, Toromanoff et al.~\cite{r18} used a similar approach of extracting affordances to effectively leverage RL for lane- keeping, collision avoidance, and traffic light detection on the CARLA simulator. They used the affordances and past vehicle speed as the RL agent's input, thereby reducing the replay buffer's memory size by 35 times, which resulted in faster training of the RL agent. They showcased outstanding results on CARLA and NoCrash benchmarks. However, it used a value-based RL approach along with discrete action spaces.

\section{Methodology}

This section describes our novel approach for learning the end-to-end autonomous driving policy using object segmentation and DRL-driven WAD agent. Figure 2 presents the proposed architecture consisting of four main parts – 3.1. CARLA to provide simulation environment, 3.2. The detection module to detect and highlight the various dynamic objects on the road, 3.3. Encoder-Decoder model to extract the latent state, 3.4. WAD agent to learn the driving policy and provide control commands to CARLA.

\subsection{Autonomous Driving Simulator}

We use the most recent CARLA 0.9.6 simulator~\cite{r16}. It provides a complex urban-like environment with multi-agent dynamics, pedestrians, intersections, cross-traffic, roundabout, and changing weather conditions, which largely differentiate it from race track simulators. Good documentation and community support makes it easy to train and validate driving policies.

With a motive of building a simulator agnostic end-to-end autonomous driving policy, we chose to utilize only four data points from the CARLA: 1. Front camera RGB images, 2. Waypoints data, 3. Traffic light state, and 4. Collision sensor data. Using these data points at each time step, we build the new transformed input state passed to the WAD agent for learning the driving policy.

\subsection{Detection Module}
The detection module is responsible for detecting the critical information present in the front camera RGB image. It mainly does three things viz detection of objects on the road, lane detection, and traffic light state detection

\textbf{Object Detection}: We used an object detection technique for detecting objects on the road using the pre-trained CenterNet ResNet50 V2 512x512 model from TensorFlow 2.0 object Detection API~\cite{r19}. We modified it to detect eight different objects on the front camera image – car, motorcycle, bicycle, person, bus, truck, traffic light, and bench.

\begin{figure}[t]

   \includegraphics[width=1.0\linewidth]{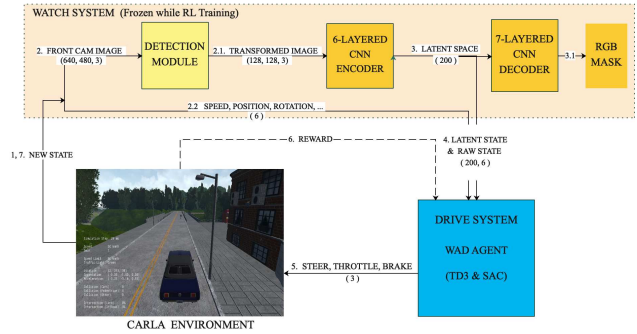}

   \caption{Proposed Architecture. The CARLA environment provides the new state, which passes through the detection module to detect objects. Detection module provides the transformed and rescaled image to the encoder-decoder model for latent state extraction. The extracted latent state and raw state inputs are passed to WAD agents to provide control commands to CARLA.}
\label{fig:long}
\label{fig:onecol}
\end{figure}

Because of the model's low latency and a detection speed of 27ms on moving frames, we maintained healthy frames per second rate. Once the objects are detected, we color the detected objects to provide the object segmentation. We preferred this object segmentation technique over semantic segmentation, to preserve the weather and road conditions. Also, performing this let the driving agent know that it has to avoid collision with specific coloured objects. To achieve 100\% collision avoidance, it does not matter to the driving agent whether the object is a car, a motorcycle, or a pedestrian. All it needs to know is, it has to avoid collision with specific-coloured objects. Using this technique, encoder-decoder model extracted and passed more meaningful latent information to the driving agent, which helped in agent's faster learning.

\textbf{Lane Detection}: It is crucial to let the driving agent know the drivable surface and avoid the risk of drifting off the driving lane or getting into other lanes. This detection technique can further help in avoiding collisions from vehicles in other lanes. We used the OpenCV~\cite{r28} library to extract the region of interest from the image. We then applied the canny edge detection technique, followed by thresholding and hough line transformation, to detect lanes in the image. However, this lane detection technique does not work on cross intersections, as there are no lanes to detect.

\textbf{Traffic Light Detection}: The driving policy needs to learn traffic-light states. To stop on a red signal and move only on a green traffic light signal. Various pre-trained deep learning models like YOLOv4~\cite{bochkovskiy2020yolov4}, TensorFlow 2.0 Object Detection API~\cite{r19} can be re-trained on a custom dataset to identify the traffic light state. However, due to computational resource constraints, we decided to use the traffic light state from the CARLA itself.

Once the objects, lanes, and traffic light state are detected and highlighted on the same image, the transformed image is then resized to a smaller size (128 x 128 x 3). The idea behind resizing the image to smaller scales is that since the critical information is already highlighted, there is minimal risk of losing out on valuable information after compression. Figure 3 shows vehicles and pedestrians coloured blue, lane detection in green and traffic light state coloured red only when the agent is on the intersection in the red-light state.

\begin{figure}[b]

   \includegraphics[width=1.0\linewidth]{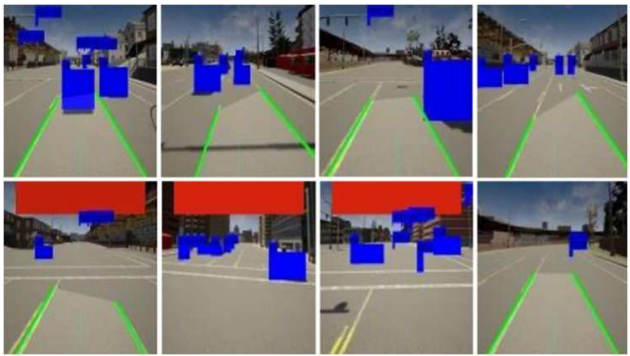}

   \caption{Samples of the encoded and transformed images.}
\label{fig:long}
\label{fig:onecol}
\end{figure}

\subsection{Improved Latent State}

Even after the RGB image is transformed and rescaled, it is still high dimensional. We use a custom encoder-decoder model to extract latent information from the transformed image. The WAD agent could easily utilize this information for learning the effective driving policy. This approach has couple of significant advantages: 1. It helps in dimensionality reduction and extraction of low- level feature information, 2. The WAD agents will have fewer parameters as the input size is drastically reduced. Thus, with reduced memory requirements and fewer parameters to train, learning becomes faster.

\subsection{Watch and Drive Agents}

The latent state along with raw states are provided as inputs to the DRL-driven WAD agent. The raw state input consists of six different parameters - speed, lateral distance, road option from next nearest waypoint, rotation, vehicle in front, and traffic light state. The latent information alone could have been provided as input to RL Agent. However, the consequence of such an approach
would be: • Only one source of truth to the RL agent as
compared to currently two. • Encoder-decoder size would
have been large. It could not have been compressed as the possibility of losing critical information increases with lower-dimensional images, leading to more RAM, GPU utilization, and longer training time. • Additionally, most of the raw states can be available with processing in real-world driving scenarios. So, including them had an added advantage.

The final input state thus formed is then passed to the WAD agent to predict three continuous steering, acceleration, and braking actions. We chose to use DRL based actor-critic methods as they combine the benefits of temporal difference and q-learning techniques used in value-based methods~\cite{r24, r25} along with directly learning continuous policy as in policy-based methods~\cite{r26,r27,r31}. The ''actor'' updates the policy distribution and is used to select actions. The ''critic'' estimates the value function and plays the role of critic over the actions predicted by the actor. We chose to
implement two state-of-the-art actor critics methods – twin delayed deep deterministic policy gradient (TD3)~\cite{r21} and soft actor-critic (SAC)~\cite{Haarnoja2018SoftAO} methods for learning the driving policy. WAD-T corresponds to watch and drive agent based on the TD3 method, and WAD-S corresponds to the agent based on the SAC method.

\subsection{Reward Mechanism}
The reward mechanism plays a critical role in robust training and faster convergence of the policy. Our reward mechanism considers the WAD agent's state in the context of
the CARLA environment. We use only the waypoint data, traffic light state, and collision sensor data from the CARLA to determine the reward. Our cumulative reward function is the combination of the following rewards:
\[ R = r_{timeout} + r_{collision} + r_{lane} + r_{speed} + r_{light} + r_{steps} \]

\(r_{timeout}\) is a one-time negative reward and leads to
terminal state if the driving agent does not move for some continuous timesteps. \(r_{collision}\) is also a one-time negative reward and leads to terminal state. It accounts for
the intensity of the collision instead of giving a fixed negative
reward. \(r_{lane}\) is a positive reward when the agent remains 
within the lane thresholds and is negative if it exceeds the lane threshold. The negative reward is variable, with the penalty increasing as the agent moves further away from the lane.

\(r_{speed}\) is one of the crucial rewards. Depending upon the agent's latitudinal and longitudinal speed, this reward can both be positive and negative. For latitudinal speed – the driving agent is rewarded depending upon the magnitude of steering angle to reduce oscillations and improve driving smoothness. For longitudinal speed - driving slow on an empty road, or driving fast when a vehicle ahead, or driving above the speed limits earns a negative reward. In comparison, driving within speed limits or stopping at red traffic signals, or stopping for any vehicle ahead, earns a positive reward. This mechanism gives the agent instant feedback on its speed-related actions.

\(r_{light}\) is a unique reward that can be positive or 
negative and can also mark the state as terminal and non-
terminal. If the agent crosses an intersection in a red-light
state, it earns a negative-reward proportional to its speed. And
if the agent fails to stop, then a one-time high negative reward
is given, and the state is marked as terminal. On the
other hand, if the agent stops at the red light, it starts receiving
a positive reward, letting the agent learn from its
reward. \(r_{steps}\) is the per step positive reward given to the 
agent. Higher the number of steps, higher is the reward agent receives. This way incentivizes the agent to stay alive for a longer duration. All these rewards together form the cumulative reward the WAD agent receives at the end of each timestep.

\section{Network Architectures}

\subsection{Latent State Extraction: Encoder-Decoder Model}

We train our encoder-decoder model on (100 + 50) k transformed image dataset collected by passing the raw front camera CARLA images through the detection module. We run the ego-vehicle in auto-pilot and manual modes under three different weather conditions.

\begin{figure}[t]
\begin{center}
   \includegraphics[width=1.0\linewidth]{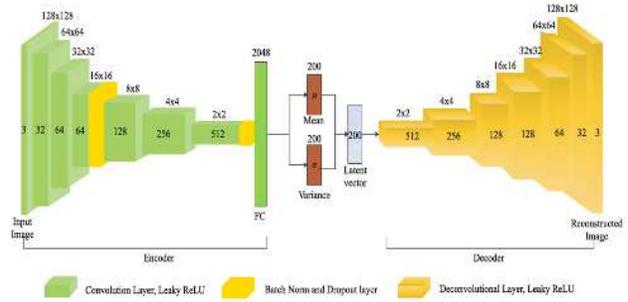}
\end{center}
   \caption{Encoder decoder model architecture}
\label{fig:long}
\label{fig:onecol}
\end{figure}

Figure 4 shows the encoder-decoder model architecture. We use the six-layered CNN as the encoder with channels ranging from 32 to 512. The kernel is set at a size of 3 x 3 and a stride of 2. L2 regularization is used with batch normalization and dropout after the second and fifth layers to overcome the overfitting problem and improve regularization. Similarly, a seven-layered CNN is used as the decoder with 3 x 3 kernel and stride of 2. The latent state space of 200 is used to extract the information. The model is trained from scratch using Adam optimizer~\cite{r30}, with a learning rate of 1e-4. Figure 5 shows the encoding results where the first row shows transformed images and reconstructed images in the second row. The model weights are frozen during the WAD agents training and evaluation.

\begin{figure}[h]
\begin{center}
   \includegraphics[width=1.0\linewidth]{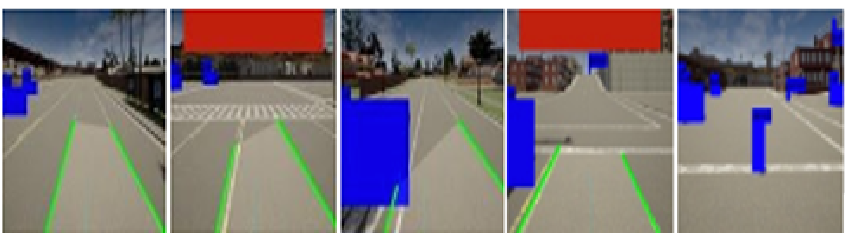}
   \includegraphics[width=1.0\linewidth]{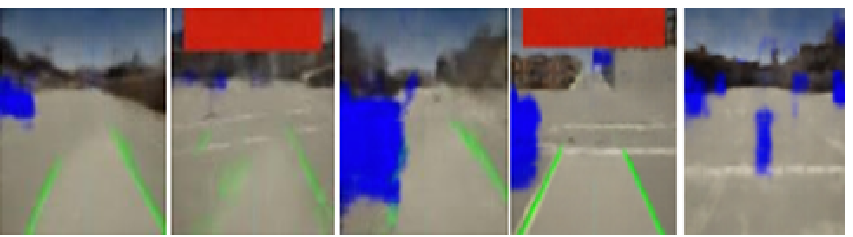}
\end{center}
   \caption{Transformed image in row 1 and corresponding decoder reconstruction in row 2}
\label{fig:long}
\label{fig:onecol}
\end{figure}

\subsection{Watch and Drive - Twin Delayed Deep Deterministic Policy Gradient (WAD-T)}

WAD agent based on the TD3 method has one policy network and two Q-networks. The policy network consists of a non-linear topology with multiple inputs and outputs. The latent state is passed through two layers of 150 and 100, whereas the raw input is passed through one hidden layer of 50. The raw input and latent information is then concatenated to pass through two linear layers of 100, 75. The output layers consist of three different actions with the first action of steering with tanh activation and other two actions of acceleration and braking with sigmoid activation. Rest all layers have ReLU activation. The critic network is slightly different from the actor network. The combined input with corresponding actions is fed as the input, and the last layer has only one output denoting the q-value. ReLU activation is followed by sigmoid activation in the last layer. All networks are trained with Adam optimizer with a learning rate of 3e-4 and 3e-3, respectively. Figure 6 describes TD3's actor network on left and critic network on right.

\begin{figure}[t]
\begin{center}
   \includegraphics[width=1.0\linewidth]{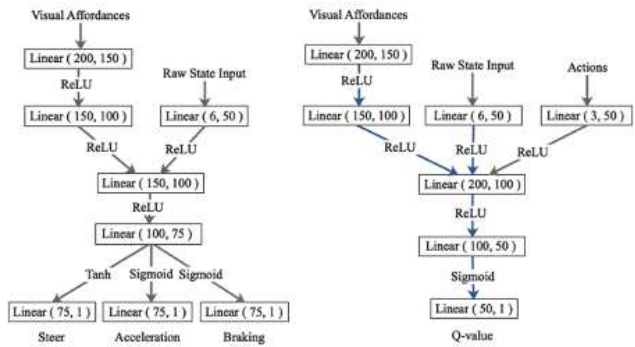}
\end{center}
   \caption{WAD-T actor network on left and critic network on right}
\label{fig:long}
\label{fig:onecol}
\end{figure}

\subsection{Watch and Drive - Soft Actor Critic (WAD-S)}

The WAD agent based on the SAC method uses one gaussian policy network, two critic networks, and one value network. The algorithm has a linear topology with latent state and raw state concatenated into one tensor and passed as an input to the model. The policy network has three hidden layers of [256, 128, 64] and a final output layer has two outputs with tanh activation. The first is the mean of the actions, and the second is the variance. The Q-network and value network have the same architecture, with the q- network taking actions as one extra input. All networks used Adam optimizer with a learning rate of 3e-4.

\section{Experiments and Results}

This section defines our set of different driving tasks and metrics for step-by-step learning of full autonomous driving. Then we thoroughly evaluate our two agents – WAD-T and WAD-S on these metrics to make fair comparisons. Later, we compare the results of our driving policies with state-of-the- art policies on the famous CARLA (CoRL2017) benchmark and further complex NoCrash benchmark. Note: All models are developed, trained, and evaluated on Windows 10 OS under python development environment utilizing only 16 GB RAM and 4 GB NVIDIA GeForce GTX 1650 GPU with CUDA 10.

\subsection{Driving Tasks for Evaluation and Comparison}
\begin{figure}[t]
\begin{center}
   \includegraphics[width=1.0\linewidth, height=0.11\textheight]{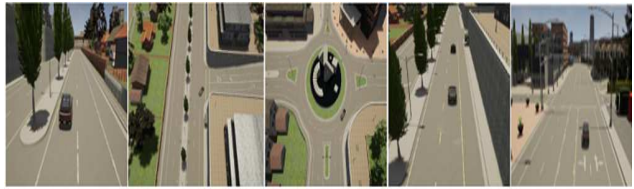}

   \caption{Five different driving tasks shown in bird view. From left to right – straight, one turn, roundabout, obstacle ahead, cross intersection.}
\label{fig:long}
\label{fig:onecol}
\end{center}
\end{figure}

We define the five driving tasks that we would be using to train and test. These driving tasks are based on their increasing difficulty as shown in figure 7 are: • Straight: The destination is 400 m away on a straight and empty road.
• One turn: The destination is one turn away from the starting point. Distance to travel: 400 m.
• Roundabout: The RL agent needs to drive through the roundabout, entering from the first exit, and exiting the roundabout from the third exit.
• Obstacle ahead: There is an object placed in the lane of the agent. The agent needs to stop at an optimal distance from the object and avoid the collision.
• Cross Intersection: The agent needs to drive through the cross intersection following traffic signals. Distance to travel: 500 m.

We evaluate the agents in town 3 and town 5 of CARLA (0.9.6) and under five different weather conditions. We use clear day, cloudy daytime and daytime rain weather conditions for training and testing under wet and cloudy sunset and mid rainy at daytime weather conditions. 



The evaluation for each task is carried out over a combination of a task, town and weather conditions. Each combination runs over 25 episodes. The episode is considered successful if the agent can drive through the set timesteps limit or reach the destination without colliding with any other object or jumping a red light, or going out of a lane. We keep a strict episode termination policy. The episode terminates if the agent goes out of lane beyond the set threshold limit, goes to the sidewalk, jumps the red-light, or collides with any other object with intensity above a fixed threshold. More the number of successful episodes, better the agent.

\subsection{Evaluation Results on Five Driving Tasks}

\begin{table}[b]
\begin{center}
\begin{tabular}{ | p{1.6cm} | p{1.2cm} | p{1.2cm} | p{1.2cm} | p{1.2cm} |}
\hline
 & \multicolumn{2}{|c|}{Training conditions} & \multicolumn{2}{|c|}{New town \& weather} \\
\hline
Task & WAD-S & WAD-T & WAD-S & WAD-T \\
\hline
Straight              & 24 ± 0.8       & \textbf{25} ± 0.0      & 24 ± 1.0       & \textbf{25} ± 0.0      \\
\hline
One turn              & 24 ± 0.0       & \textbf{25} ± 0.0      & 23 ± 1.0       & \textbf{25} ± 0.0      \\
\hline
Roundabout            & 23 ± 0.4       & \textbf{24} ± 0.4      & 23 ± 0.5       & \textbf{24} ± 0.5      \\
\hline
Obstacle ahead        & 23 ± 0.4       & \textbf{25} ± 0.0      & 22 ± 1.0       & \textbf{24} ± 1.0      \\
\hline
Cross intersection    & 23 ± 1.2      & \textbf{24} ± 0.4      & 22 ± 1.0       & \textbf{24} ± 0.5 \\
\hline
\end{tabular}
\end{center}
\caption{Mean and standard deviation of WAD agents under train and test conditions.}
\end{table}

We assessed our two agents on five separate tasks under the train and test conditions. Note that we use the same agents and do not fine-tune them separately for each scenario. Table 1. presents the mean and standard deviation of the two agents on five driving tasks in train and test environments under varying weather conditions. WAD-T and WAD-S agents performed well under train and test conditions, with WAD-T having a slightly better mean and less deviation on all five tasks.

\subsection{Evaluation Results on Composite Driving Task}

We also evaluated the agents on composite task under dense traffic. There were 70 vehicles, and 100 pedestrians spawned at random locations throughout the town and set to move in autopilot mode. The WAD agent's vehicle was spawned each time at a random location. We performed this evaluation in both the train and test environment settings with 100 episodes each. Figure 8 shows the cause of termination of the WAD agent episodes in train and test settings, respectively.

\begin{figure}[h]
\begin{center}
   \includegraphics[width=1.0\linewidth]{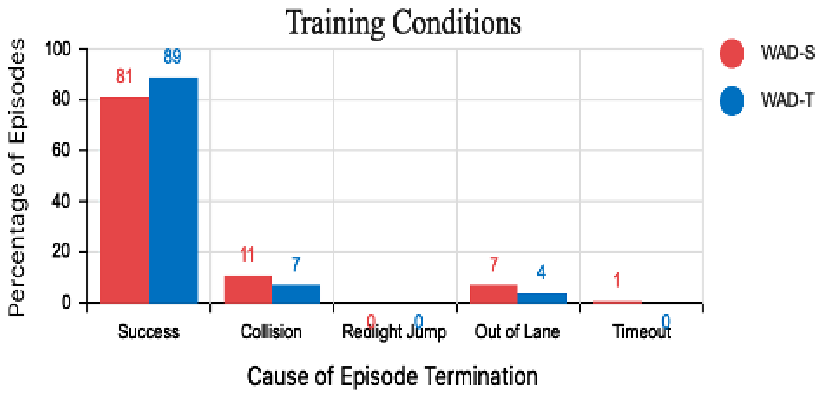}
   \includegraphics[width=1.0\linewidth]{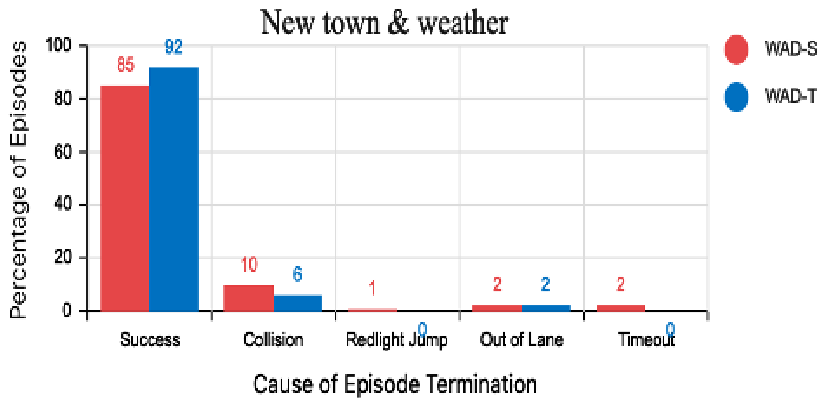}
\end{center}
   \caption{Cause of termination of episodes under train and test condition}
\label{fig:long}
\label{fig:onecol}
\end{figure}

Both the agents perform significantly well, with 92\% successfully completed episodes in WAD-T and 85\% in WAD-S under new town and weather conditions. The WAD agents can follow the lane, stop and slow down at the cross intersection, maintain the proper distance from other vehicles, and avoid a collision. Collision percent for WAD-T is 6-7\%, whereas it is around 10-11\% for WAD-S. Interestingly, both the agents have learned to stop at the red light with only 0-1\% of episode termination due to red light jumps. The WAD-T agent outperforms the other with a higher success rate and fewer collisions under both train and test conditions.

\subsection{CARLA and NoCrash Benchmarks}

We performed the comparison of our WAD agents on the original CARLA benchmark~\cite{r16} (CoRL2017) and the recent NoCrash benchmark~\cite{r6}.

\textbf{CARLA (CoRL2017) Benchmark}: The CoRL2017 benchmark consists of four driving tasks of straight, one turn, navigation, navigation with dynamic objects. We choose Town 3 for training and Town 5 for testing. We have considered five weather conditions of clear day, wet cloudy sunset and daytime rain for training and cloudy daytime and hard rain at daytime for testing purposes.

\begin{table*}
\begin{center}
\begin{tabular}{ | p{2.0cm} | p{0.6cm} | p{0.6cm} | p{0.8cm} | p{0.7cm} | p{0.7cm} | p{0.7cm} | p{0.7cm} | p{0.6cm} | p{0.6cm} | p{0.8cm} | p{0.7cm} | p{0.7cm} | p{0.7cm} | p{0.7cm} |}
\hline
 CoRL 2017 & \multicolumn{7}{|c|}{Training conditions} & \multicolumn{7}{|c|}{New town \& weather} \\
 \hline
 Task & RL & CAL & CILRS & LBC & IA & WAD-S & WAD-T & RL & CAL & CIRLS & LBC & IA & WAD-S & WAD-T \\
 \hline
 Straight & 89 & 100 & 96 & 100 & 100 & 100 & 100 & 68 & 94 & 96 & 100 & 100 & 100 & 100\\
 One turn & 34 & 97 & 92 & 100 & 100 & 100 & 100 & 20 & 72 & 84 & 100 & 100 & 100 & 100\\
 Navigation & 14 & 92 & 95 & 100 & 100 & 96 & 100 & 6 & 68 & 69 & 98 & 100 & 96 & 100\\
 Nav. dynamic & 7 & 83 & 92 & 100 & 100 & 96 & 100 & 4 & 64 & 66 & 99 & 98 & 98 & 100\\
 \hline
\end{tabular}
\end{center}
\caption{Quantitative evaluation of our two autonomous driving agents WAD-S and WAD-T on goal-directed navigation tasks on CARLA (CoRL2017) benchmark. The table presents the success rate of models to the state-of-the-art RL~\cite{r16}, CAL~\cite{r23}, CILRS~\cite{r6}, LBC~\cite{r3}, IA~\cite{r18} models with varying train and test conditions on CARLA 0.9.6.}
\end{table*}

\begin{table*}
\begin{center}
\begin{tabular}{ | p{2.2cm} | p{0.8cm} | p{0.8cm} | p{0.8cm} | p{1.2cm} | p{1.2cm} | p{0.8cm} | p{0.8cm} | p{0.8cm} | p{1.2cm} | p{1.2cm} |}
\hline
NoCrash & \multicolumn{5}{|c|}{Training conditions} & \multicolumn{5}{|c|}{New town \& weather} \\
 \hline
 Task & CILRS & LBC & IA & WAD-S & WAD-T & CIRLS & LBC & IA & WAD-S & WAD-T \\
 \hline
 Empty & 97 & 100 & 100 & 98 & 100 & 90 & 70 & 99 & 96 & 100 \\
 Regular Traffic & 97 & 100 & 100 & 98 & 100 & 90 & 70 & 99 & 96 & 100 \\
 Dense Traffic & 97 & 100 & 100 & 98 & 100 & 90 & 70 & 99 & 96 & 100 \\
 \hline
\end{tabular}
\end{center}
\caption{Comparison of success rate of the presented WAD agents (WAD-S and WAD-T) with state-of-the-art baselines CILRS~\cite{r6}, LBC~\cite{r3}, IA~\cite{r18} on Nocrash benchmark under varying train and test conditions on CARLA 0.9.6}
\end{table*}

We evaluate the models on 25 episodes for each combination of task, town, and weather. The episode is considered successful if the driving agent can reach the destination within the time limit. The destination is set far enough such that the agent can reach it with a speed of 10km/hr. The benchmark ignores traffic light violations. And infractions such as driving on sidewalks or collisions do not lead to termination of the episode.

We compare our agents with five state of the art autonomous driving agents, that is 1. Reinforcement learning (RL) agent in CARLA~\cite{r16}, 2. Conditional affordance learning (CAL) agent~\cite{r23}, 3. CILRS~\cite{r6}, 4. Learn by cheating (LBC)~\cite{r3}, 5. Implicit affordances (IA)~\cite{r18}. Table 2 presents the comparison of the success rate of our proposed agents to the state-of-the-art baselines on the original CARLA benchmark (CoRL2017) under both train and test conditions. The results show that our WAD agents achieve a 100\% success rate on the CoRL2017 benchmark. The agents outperform baseline models by a significant margin and have comparable results with LBC and IA.

\textbf{NoCrash Benchmark}: The recent and complex NoCrash benchmark~\cite{r6} consists of three different tasks: - 1. Empty town, 2. Regular traffic, and 3. Dense traffic with a large number of vehicles and pedestrians. We use town3 and town5 for training and testing, respectively, under four weather conditions. The episode is considered successful if the agent can reach the destination without colliding with any object. We terminate the episode when the collision above fixed magnitude happens, and the episode is not terminated on the traffic light violation. Instead, it is noted, and the episode continues.
We compare our proposed agents with three state-of-the-art autonomous driving agents - CILRS~\cite{r6}, learn by cheating (LBC)~\cite{r3}, and implicit affordances (IA)~\cite{r18}. Table 3 presents the comparison of the success rate of our proposed agents to state-of-the-art baselines on the NoCrash benchmark under train and test settings.
Our proposed WAD-T agent achieves 82\% success rate on dense traffic scenario of NoCrash benchmark in test town 5 under new weather conditions outperforming CILRS~\cite{r6}, LBC~\cite{r3}, and IA~\cite{r18} models by a significant margin. The agent improves state-of-the-art by +16\% under train conditions and by +40\% under test conditions on the NoCrash benchmark.

\subsection{Infraction Analysis}

\textbf{Average Distance, Speed, Time Alive on Composite Driving Task} We tried to evaluate further and compare our agents based on average distance travelled, average speed, top speed, time alive, and the reward they received. This experiment truly demonstrates the quality of trained agents as, unlike other experiments, we had 4000 timesteps to monitor over 100 episodes under test conditions. This effectively meant driving in the town for about 11.11 minutes as we were having a frame rate of six per second. Also, the driving agent would come across multiple cross- intersections, roundabout, moving traffic, pedestrians, and other objects, all in a single episode. We terminated the episode when the agent went off the road, collided with any object, or jumped the signal. Table 4 shows the comparison of two agents on average distance, time alive, average speed and top speed.

\begin{table}[h]
\begin{center}
\begin{tabular}{ | p{3.4cm} | p{1.8cm} | p{1.8cm} | }
\hline
 Parameters & WAD-S & WAD-T \\ 
 \hline
 Avg. distance travelled & 1.06 kms & 2.15 kms \\  
 Avg. time alive & 4.94 min & 7.39 min  \\ 
 Avg. speed & 13.32 km/hr & 17.49 km/hr  \\
 Top speed  & 17.28 km/hr & 26.42 km/hr  \\ 
 \hline
\end{tabular}
\end{center}
\caption{WAD agent's comparison on avg distance, speed, time alive under new town and weather conditions}
\end{table}

On average, the WAD-S agent was able to drive appropriately at a speed of 13.32 km/hr for about 1791.97 timesteps (~ 4.94 minutes), covering an average distance of 1.06 kms and receiving an average reward of 9807.02 over the episodes. In contrast, the WAD-T agent was able to stay alive for 2662.80 timesteps (~7.39 minutes), with an average speed of 17.49 km/hr, covering an average distance of 2.15 kms and receiving an average reward of 15519.32. WAD-S agent achieves a top speed of 17.28 km/hr, whereas WAD-T agent reaches a maximum speed of 26.42 km/hr.




\section{Conclusion and Future Work}

In this work, we introduced two Watch and Drive agents (WAD-T, WAD-S) for learning the optimal autonomous driving policy in the CARLA simulation using object segmentation, latent state extraction and reinforcement learning. We trained and evaluated the agents on different driving tasks under varying train and test conditions. We also compared our agents with state-of-the-art agents on the CARLA and NoCrash benchmarks. The results showed that the WAD agents achieved remarkable results on the CoRL2017 and NoCrash benchmarks outperforming baseline models across techniques. Three important factors that led us to achieve good results – 1. Step-by-step learning of different driving tasks and then training over the composite driving task, 2. Hard episode termination policy, 3. Dense rewarding mechanism that took into consideration almost every possible scenario. Additionally, we achieved state-of-the-art results using very low computational resources.  

In future work, it would be interesting to apply deep learning based lane detection, traffic signs and speed limit detection algorithms for intelligent decision-making. Also, inspired by the work done by Pan et al.~\cite{r37}, we plan to extend our study and experiment with our WAD agents under real-world environments.

{\small
\bibliographystyle{ieee_fullname}
\bibliography{main}
}

\end{document}